\documentclass[11pt]{article}
\usepackage{CJKutf8}

\usepackage{graphicx}
\usepackage{booktabs} 
\usepackage{makecell}
\usepackage{amsfonts,amssymb} 
\usepackage{float}

\usepackage{acl}
\usepackage{times}
\usepackage{latexsym}

\usepackage[T1]{fontenc}

\usepackage[utf8]{inputenc}

\usepackage{microtype}
\title{MiniRBT: A Two-stage Distilled Small Chinese Pre-trained Model}

\author{
Xin Yao$^\dag$,
Ziqing Yang$^\dag$,
Yiming Cui$^\ddag$$^\dag$,
Shijin Wang$^\dag$ \\
{$^\dag$State Key Laboratory of Cognitive Intelligence, iFLYTEK Research, Beijing, China} \\
{$^\ddag$Research Center for SCIR, Harbin Institute of Technology, Harbin, China} \\
$^\dag$\tt\{xinyao10,zqyang5,ymcui,sjwang3\}@iflytek.com \\
$^\ddag$\tt ymcui@ir.hit.edu.cn}

\begin{document}
\maketitle
\begin{abstract}
In natural language processing, pre-trained language models have become essential infrastructures. However, these models often suffer from issues such as large size, long inference time, and challenging deployment. Moreover, most mainstream pre-trained models focus on English, and there are insufficient studies on small Chinese pre-trained models. In this paper, we introduce MiniRBT, a small Chinese pre-trained model that aims to advance research in Chinese natural language processing. MiniRBT employs a narrow and deep student model and incorporates whole word masking and two-stage distillation during pre-training to make it well-suited for most downstream tasks. Our experiments on machine reading comprehension and text classification tasks reveal that MiniRBT achieves 94\% performance relative to RoBERTa, while providing a 6.8x speedup, demonstrating its effectiveness and efficiency.
\end{abstract}

\section{Introduction}
In recent years, the pre-trained language model based on Transformers~\cite{vaswani2017attention} has become a paradigm of natural language processing, and their performance has been increasing with the growth in model size. These models have dominated the lists of primary AI models for natural language processing and computer vision, including BERT~\cite{devlin-etal-2019-bert}, XLNet~\cite{yang2019xlnet}, RoBERTa~\cite{liu2019roberta}, SpanBERT~\cite{joshi-etal-2020-spanbert}, ELECTRA~\cite{clark2020electra}. Despite their significant progress, pre-trained models still face considerable challenges in practical applications, such as high training costs and high latencies. Furthermore, while research on pre-trained models has mainly focused on the English language, there is a lack of smaller pre-trained models for Chinese.

\begin{table*}[h]
\centering
\resizebox{\textwidth}{!}{
\begin{tabular}{ll}
\toprule
\textbf{Masking Strategy} & \textbf{Example}\\
\midrule
Original text & \begin{CJK}{UTF8}{gbsn}使用语言模型来预测下一个词的probability。\end{CJK} \\
Word segmentation & \begin{CJK}{UTF8}{gbsn}使用 \ 语言 \ 模型 \ 来 \ 预测 \ 下 一个 \ 词 \ 的 \ probability 。\end{CJK} \\
Original masking & \begin{CJK}{UTF8}{gbsn}使 \ 用 \ 语 \ 言\  [MASK] \ 型 \ 来 \ [MASK] \ 测 \ 下 \ 一 \ 个 \ 词 \ 的 \ pro [MASK] \#\#lity 。\end{CJK} \\ 
WWM & \begin{CJK}{UTF8}{gbsn}使 \ 用 \ 语 \ 言 \ [MASK] [MASK] \ 来 \ [MASK] [MASK] \ 下 \ 一 \ 个 \ 词 \ 的 \ [MASK] [MASK] [MASK] 。\end{CJK} \\ 
\bottomrule
\end{tabular}}
\caption{Examples of different masking strategies.}
\label{tab:wwm}
\end{table*}

\begin{figure*}[h]
    \centering
    \includegraphics[width=0.9\textwidth]{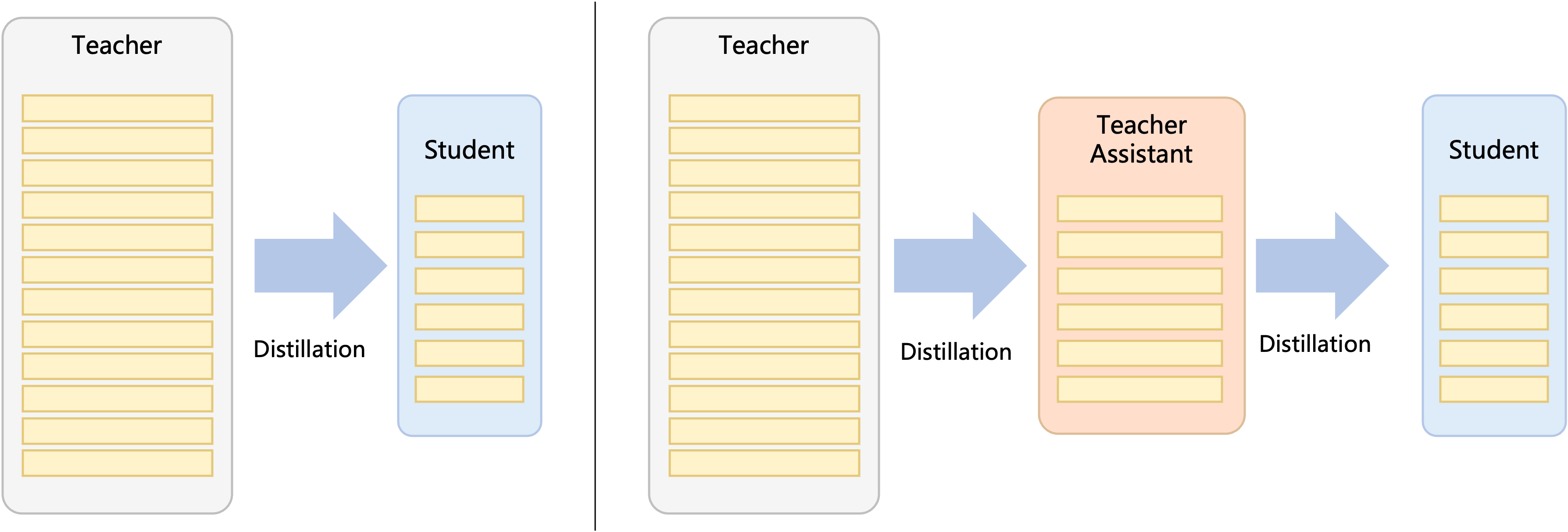}
    \caption{Comparison of one-stage distillation (left) and two-stage distillation (right) processes.}
    \label{fig:two_stage distill}
\end{figure*}

Therefore, to further advance the development of Chinese natural language processing, we propose a small Chinese pre-trained model with strong practicability. Our approach involves utilizing dynamic whole word masking techniques~\cite{cui-etal-2021-pretrain} to generate training samples that facilitate comprehensive modeling of coarse-grained semantics. We apply a narrow and deep model structure for the student model. During the pre-training, we employ a two-stage distillation method that incorporates a teacher assistant model. We first distill from the teacher to the teacher assistant, and then from the teacher assistant to the student. This results in a lighter and faster model that can be fine-tuned for multiple downstream tasks, demonstrating excellent performance.

Our research focuses primarily on developing small Chinese pre-trained models that can be applied to practical tasks. The main contributions of this paper can be summarized as follows:

\begin{enumerate}

\item We conduct experiments that conclude that a narrower and deeper network structure is more effective than a wide and shallow structure of similar size.

\item Based on the above finding, we propose MiniRBT\footnote{Available
at~\url{https://github.com/iflytek/MiniRBT}}, a small pre-trained model for Chinese which is only 10\% the size of Chinese RoBERTa, while maintaining an average performance of 94\% compared to RoBERTa. Moreover, it offers a 6x-7x speedup.

\end{enumerate}

\section{Background}
\subsection{Chinese RoBERTa-wwm}

Since the emergence of BERT, pre-trained language models have advanced significantly and rapidly. However, conventional Chinese pre-trained models, such as the original Chinese BERT, typically employ a segmentation method that divides Chinese sequences at the granularity of individual characters, which means that Chinese words are segmented into characters using a WordPiece-based word segmentation approach. These characters are randomly masked individually during training sample generation without considering Chinese word segmentation. To address this issue, Chinese RoBERTa-wwm~\cite{cui-etal-2021-pretrain} uses a whole word masking (WWM) method specifically designed for Chinese. When part of a whole Chinese word is masked, other parts of the same word are also masked. It should be noted that the WWM method only affects the selection of masking tokens during the pre-training stage. Chinese RoBERTa-wwm is trained on Chinese Wikipedia (both Simplified and Traditional). In our work, we utilize Chinese RoBERTa-wwm as our teacher model.

\subsection{Knowledge Distillation}

In recent years, a growing number of works on model compression have been proposed to reduce the number of model parameters and improve the speed of model inference, such as quantization~\cite{shen2020q, fan2020training}, pruning~\cite{xia2022structured, lagunas-etal-2021-block}, knowledge distillation~\cite{jiao-etal-2020-tinybert, sun-etal-2020-mobilebert, distillbert,hou2020dynabert}. Knowledge distillation (KD) transfers the knowledge embedded in a large teacher model to a small student model by training the student to mimic the behaviors of the teacher. For instance, TinyBERT~\cite{jiao-etal-2020-tinybert} first obtains a general distilled small model by performing general distillation on the large-scale corpus from a general domain, and then performs task-specific distillation with downstream data during the fine-tuning stage. DynaBERT~\cite{hou2020dynabert} trains a width-adaptive and depth-adaptive BERT by distilling knowledge from full-sized models to small sub-networks. KD is also applied to pruning to enhance performance, such as block pruning~\cite{lagunas-etal-2021-block} and CoFi~\cite{xia2022structured}.

\section{Method}
In this section, we introduce the relevant methodologies employed by MiniRBT.

\subsection{Whole Word Masking}
The original WordPiece-based word segmentation method separates Chinese input sequences into independent characters and randomly masks the characters. This method simplifies prediction since the model can only remember particular character orders in words to predict masked characters without considering the semantic contextual relationships. To enhance the model's performance, we adopted the dynamic Whole Word Masking (WWM) approach as a training sample generation strategy during the pre-training. We begin with utilizing the conventional Chinese word segmentation (CWS) method to segment the input into Chinese words, and are subsequently masked using WWM, as presented in Table~\ref{tab:wwm}. We employed LTP~\cite{che2010ltp} as our preferred tool to extract word segmentation information.
\begin{table*}[ht]
\centering
\resizebox{\textwidth}{!}{
\begin{tabular}{lccccc cc}
\toprule
\textbf{Model} & \textbf{Layers} & \textbf{Hidden Size} & \textbf{FFN Size} & \textbf{Heads}  & 	\textbf{Model Size} & \textbf{Model Size (w/o embeddings)} & \textbf{Speedup}\\
\midrule
RoBERTa-wwm            & 12 & 768  & 3072  & 12    & 102.3M (100\%)  & 85.7M (100\%)  & 1x   \\
RBT6 (KD)             & 6  & 768  & 3072  & 12    & 59.8M (58.4\%) & 43.1M (50.3\%) & 1.7x \\
RBT3                  & 3  & 768  & 3072  & 12    & 38.5M (37.6\%)  & 21.9M (25.6\%) & 2.8x \\
\textbf{RBT4-H312}    & 4  & 312  & 1200  & 12    & 11.4M (11.1\%)  & 4.7M (5.5\%) & 6.8x \\
\textbf{MiniRBT-H256} & 6  & 256  & 1024  & 8     & 10.4M (10.2\%)  & 4.8M (5.6\%)  & 6.8x\\
\textbf{MiniRBT-H288} & 6  & 288  & 1152  & 8     & 12.3M (12.0\%)  & 6.1M (7.1\%) & 5.7x\\
\bottomrule
\end{tabular}}
\caption{Comparison of model structures. RBT stands for RoBERTa, and RBT3 is initialized by RoBERTa's first three layers and pre-trained for 1M steps. The number of layers does not include the embedding and prediction layers.}
\label{tab: model}
\end{table*}

\subsection{Two-stage Distillation}
The traditional knowledge distillation method transfers knowledge directly from the teacher to the student. However, when there is a significant difference in the structures of the teacher and student models, this approach may result in a performance gap. To address this issue, we proposed using a two-stage distillation approach during the pre-training stage, which builds on the concept of Teacher Assistant Knowledge Distillation~\cite{mirzadeh2020improved}. As depicted in Figure~\ref{fig:two_stage distill}, this method involves distilling knowledge from the teacher (RoBERTa) to the teacher assistant (RBT6), and then from the teacher assistant to the student (MiniRBT). The intermediate step of the teacher assistant helps to reduce the size gap between the teacher and the student model, subsequently improving the student models' performance in downstream tasks.

To apply knowledge distillation (KD) with hidden layer distillation and prediction layer distillation, we employed TextBrewer \cite{yang-etal-2020-textbrewer}, a PyTorch-based model distillation toolkit designed for natural language processing. We distill the knowledge from the output of the hidden layer. The objective is
\begin{equation}
    \mathcal{L}_\mathrm{layer}  = \sum \mathrm{MSE}(H^{i'}_sW_h, H^i_t)
\end{equation}
where the matrices $ H^{i'}_s \in \mathbb{R}^{l\times d'}$ and $H^i_t \in \mathbb{R}^{l\times d} $ represent the hidden representation of the $i'$-th student's hidden layer and the i-th teacher's hidden layer respectively. The $ W_h \in \mathbb{R}^{d'\times d} $ is a linear transformation that matches the hidden state of the student network and the hidden state of the teacher network. Apart from mimicking the hidden layer behavior of the teacher, we also trained the student model by employing the cross-entropy loss with the teacher's soft target probability
\begin{equation}
\mathcal{L}_\mathrm{pred} =-p(z^T)\cdot\log p(z^S)
\end{equation}
where $z^S$ and $z^T$ are the logits vectors predicted by the student and teacher respectively, and $p=\mathrm{softmax}(z/t)$ is the scaled probability with temperature $t$ and logits $z$.

Finally, we combine the hidden layer distillation with the prediction layer distillation:
\begin{equation}
    \mathcal{L}_\mathrm{distill} = \mathcal{L}_\mathrm{layer} + \mathcal{L}_\mathrm{pred}
\end{equation}

\subsection{Narrower and Deeper Students}
Through preliminary experiments, we find that a narrow and deep model structure outperforms a wide and shallow one, when they have the same number of parameters. Hence, we employed a narrow and deep design for MiniRBT. We present the details of the model structure in Table~\ref{tab: model}. MiniRBT consists of two branches of models, MiniRBT-H256 and MiniRBT-H288. These models have hidden layer dimensions of 256 and 288, respectively, and contain 6 transformer layers, pre-trained via the two-stage distillation approach.

\begin{table*}[htb]
\centering
\resizebox{\textwidth}{!}{
\begin{tabular}{lcccccccc}
\toprule
\textbf{Task} & \textbf{\makecell{CMRC 2018 \\ (F1/EM)}}   & \textbf{\makecell{DRCD \\ (F1/EM)}}  & \textbf{\makecell{OCNLI \\ (Acc)}} & \textbf{\makecell{LCQMC \\ (Acc)}} & \textbf{\makecell{BQ Corpus \\ (Acc)}} & \textbf{\makecell{TNEWS \\ (Acc)}} & \textbf{\makecell{ChnSentiCorp \\ (Acc)}}  \\
\midrule
RoBERTa  & 87.30/68.00     & 94.40/89.40   & 76.58 & 89.07 & 85.76     & 57.66 & 94.89         \\
RBT6 (KD) & 84.40/64.30   & 91.27/84.93 & 72.83 & 88.52 & 84.54     & 55.52 & 93.42         \\
RBT3   & 80.30/57.73  & 85.87/77.63 & 69.80 & 87.30  & 84.47     & 55.39 & 93.86         \\
\textbf{RBT4-H312}    & 77.90/54.93  & 84.13/75.07 & 68.50 & 85.49 & 83.42     & 54.15 & 93.31         \\
\textbf{MiniRBT-H256} & 78.47/56.27 & 86.83/78.57 & 68.73 & 86.81 & 83.68     & 54.45 & 92.97         \\
\textbf{MiniRBT-H288} & 80.53/58.83 & 87.10/78.73  & 68.32 & 86.38 & 83.77     & 54.62 & 92.83   \\
\bottomrule
\end{tabular}}
\caption{Comparison of MiniRBT with baseline models on reading comprehension task and text classification task.}
\label{tab:results}
\end{table*}
\begin{table}
\centering

\begin{tabular}{lccc}
\toprule
\textbf{Model}  & \textbf{CMRC 2018}   & \textbf{LCQMC} \\
\midrule
two-stage & \textbf{77.97/54.60} & \textbf{86.58} \\
one-stage & 77.57/54.27  & 86.39 \\ 
\bottomrule
\end{tabular}
\caption{Comparing the results of two-stage distillation and one-stage distillation. The model is MiniRBT-H256 pre-trained with 30K steps, which is different from the published 100K step pre-trained model.}
\label{tab:two-stage results}
\end{table}
\section{Experiments}
\subsection{Downstream Tasks}
\paragraph{Machine Reading Comprehension} Machine reading comprehension (MRC) is a document-level modeling task that requires models to answer questions based on given passages. We evaluated our models on two Chinese reading comprehension datasets: CMRC 2018~\cite{cui-etal-2019-span} and DRCD~\cite{shao2018drcd}. They are similar in the form of SQuAD~\cite{rajpurkar-etal-2018-know}. The evaluation metrics are F1 and EM.
\paragraph{Text Classification} For single sentence classification, we use TNEWS and ChnSentiCorp~\cite{tan2008empirical}. The ChnSentiCorp dataset involves sentiment classification wherein texts need to be classified as either positive or negative, while TNEWS dataset involves the classification of short texts into various news categories. For sentence pair classification, we select three datasets: OCNLI, LCQMC~\cite{liu2018lcqmc}, and BQ corpus~\cite{chen-etal-2018-bq}. Both OCNLI and TNEWS are included as subtasks in the Chinese Language Understanding Evaluation (CLUE) Benchmark~\cite{xu2020clue}. The evaluation metric for these tasks is accuracy.

\subsection{Training Setup}

During the pre-training phase, a training batch size of 4096 and a peak learning rate of 4e-4 are employed, while the temperature is set to 8 and the number of training steps is 100K.

We fine-tune the model on downstream tasks for 2, 3, 5, and 10 epochs, respectively, with a learning rate chosen from $\{5e-5,1e-4\}$. To decrease the impact of randomness on the experimental results, we run each task at least three times with different random seeds and report the average performance score on the development set. It is worth noting that, for smaller models, increasing the number of iterations and the learning rate tends to improve the performance on downstream tasks.

Model speedups are evaluated relative to Chinese RoBERTa on a single NVIDIA M40 GPU. The input length for all tasks was set to 512, with a batch size of 128.

\subsection{Results}
Table~\ref{tab:results} presents the results of MiniRBT's performance on reading comprehension and text classification tasks. With only 10\% of the parameters used by Chinese RoBERTa, MiniRBT achieves over 92\% of its performance, a substantial improvement over RBT3 which has 3-4 times more parameters in reading comprehension tasks. In text classification tasks, MiniRBT achieves 98\% of RoBERTa's performance, with an average relative performance of 95.3\% compared to Chinese RoBERTa. Table~\ref{tab:results} further indicates that, with the same number of parameters (excluding the embedding layer), MiniRBT outperforms RBT4-H312, demonstrating that a narrow and deep model structure yields superior performance when compared to a wide and shallow model structure.

Table~\ref{tab:two-stage results} reveals that two-stage pre-training distillation outperforms one-stage pre-training distillation in both reading comprehension and text classification tasks. These results suggest that the two-stage distillation approach effectively reduces the gap in size between teacher and student models, thus allowing students to maintain excellent performance even with small sizes.
\section{Conclusion}
In this study, we introduce MiniRBT, a small Chinese pre-trained model pre-trained with dynamic WWM, two-stage distillation, and a narrow and deep model structure. With only 10\% of the parameters of RoBERTa, MiniRBT achieves an average relative performance of 94\% and a speedup of 6.8x. Our findings demonstrate that MiniRBT has notable performance advantages over other models with the same number of parameters and can even surpass larger models with 3-4 times more parameters. In the future, we expect to combine pruning and quantization to propose more lightweight models.
\bibliography{custom}

\begin{thebibliography}{25}
\expandafter\ifx\csname natexlab\endcsname\relax\def\natexlab#1{#1}\fi

\bibitem[{Che et~al.(2010)Che, Li, and Liu}]{che2010ltp}
Wanxiang Che, Zhenghua Li, and Ting Liu. 2010.
\newblock Ltp: A chinese language technology platform.
\newblock In \emph{Coling 2010: Demonstrations}, pages 13--16.

\bibitem[{Chen et~al.(2018)Chen, Chen, Liu, Yang, Lu, and
  Tang}]{chen-etal-2018-bq}
Jing Chen, Qingcai Chen, Xin Liu, Haijun Yang, Daohe Lu, and Buzhou Tang. 2018.
\newblock \href {https://doi.org/10.18653/v1/D18-1536} {The {BQ} corpus: A
  large-scale domain-specific {C}hinese corpus for sentence semantic
  equivalence identification}.
\newblock In \emph{Proceedings of the 2018 Conference on Empirical Methods in
  Natural Language Processing}, pages 4946--4951, Brussels, Belgium.
  Association for Computational Linguistics.

\bibitem[{Clark et~al.(2020)Clark, Luong, Le, and Manning}]{clark2020electra}
Kevin Clark, Minh-Thang Luong, Quoc~V Le, and Christopher~D Manning. 2020.
\newblock Electra: Pre-training text encoders as discriminators rather than
  generators.
\newblock \emph{arXiv preprint arXiv:2003.10555}.

\bibitem[{Cui et~al.(2021)Cui, Che, Liu, Qin, and
  Yang}]{cui-etal-2021-pretrain}
Yiming Cui, Wanxiang Che, Ting Liu, Bing Qin, and Ziqing Yang. 2021.
\newblock \href {https://doi.org/10.1109/TASLP.2021.3124365} {Pre-training with
  whole word masking for chinese bert}.

\bibitem[{Cui et~al.(2019)Cui, Liu, Che, Xiao, Chen, Ma, Wang, and
  Hu}]{cui-etal-2019-span}
Yiming Cui, Ting Liu, Wanxiang Che, Li~Xiao, Zhipeng Chen, Wentao Ma, Shijin
  Wang, and Guoping Hu. 2019.
\newblock \href {https://doi.org/10.18653/v1/D19-1600} {A span-extraction
  dataset for {C}hinese machine reading comprehension}.
\newblock In \emph{Proceedings of the 2019 Conference on Empirical Methods in
  Natural Language Processing and the 9th International Joint Conference on
  Natural Language Processing (EMNLP-IJCNLP)}, pages 5883--5889, Hong Kong,
  China. Association for Computational Linguistics.

\bibitem[{Devlin et~al.(2019)Devlin, Chang, Lee, and
  Toutanova}]{devlin-etal-2019-bert}
Jacob Devlin, Ming-Wei Chang, Kenton Lee, and Kristina Toutanova. 2019.
\newblock \href {https://doi.org/10.18653/v1/N19-1423} {{BERT}: Pre-training of
  deep bidirectional transformers for language understanding}.
\newblock In \emph{Proceedings of the 2019 Conference of the North {A}merican
  Chapter of the Association for Computational Linguistics: Human Language
  Technologies, Volume 1 (Long and Short Papers)}, pages 4171--4186,
  Minneapolis, Minnesota. Association for Computational Linguistics.

\bibitem[{Fan et~al.(2020)Fan, Stock, Graham, Grave, Gribonval, Jegou, and
  Joulin}]{fan2020training}
Angela Fan, Pierre Stock, Benjamin Graham, Edouard Grave, R{\'e}mi Gribonval,
  Herve Jegou, and Armand Joulin. 2020.
\newblock Training with quantization noise for extreme model compression.
\newblock \emph{arXiv preprint arXiv:2004.07320}.

\bibitem[{Hou et~al.(2020)Hou, Huang, Shang, Jiang, Chen, and
  Liu}]{hou2020dynabert}
Lu~Hou, Zhiqi Huang, Lifeng Shang, Xin Jiang, Xiao Chen, and Qun Liu. 2020.
\newblock Dynabert: Dynamic bert with adaptive width and depth.
\newblock \emph{Advances in Neural Information Processing Systems},
  33:9782--9793.

\bibitem[{Jiao et~al.(2020)Jiao, Yin, Shang, Jiang, Chen, Li, Wang, and
  Liu}]{jiao-etal-2020-tinybert}
Xiaoqi Jiao, Yichun Yin, Lifeng Shang, Xin Jiang, Xiao Chen, Linlin Li, Fang
  Wang, and Qun Liu. 2020.
\newblock \href {https://doi.org/10.18653/v1/2020.findings-emnlp.372}
  {{T}iny{BERT}: Distilling {BERT} for natural language understanding}.
\newblock In \emph{Findings of the Association for Computational Linguistics:
  EMNLP 2020}, pages 4163--4174, Online. Association for Computational
  Linguistics.

\bibitem[{Joshi et~al.(2020)Joshi, Chen, Liu, Weld, Zettlemoyer, and
  Levy}]{joshi-etal-2020-spanbert}
Mandar Joshi, Danqi Chen, Yinhan Liu, Daniel~S. Weld, Luke Zettlemoyer, and
  Omer Levy. 2020.
\newblock \href {https://doi.org/10.1162/tacl_a_00300} {{S}pan{BERT}: Improving
  pre-training by representing and predicting spans}.
\newblock \emph{Transactions of the Association for Computational Linguistics},
  8:64--77.

\bibitem[{Lagunas et~al.(2021)Lagunas, Charlaix, Sanh, and
  Rush}]{lagunas-etal-2021-block}
Fran{\c{c}}ois Lagunas, Ella Charlaix, Victor Sanh, and Alexander Rush. 2021.
\newblock \href {https://doi.org/10.18653/v1/2021.emnlp-main.829} {Block
  pruning for faster transformers}.
\newblock In \emph{Proceedings of the 2021 Conference on Empirical Methods in
  Natural Language Processing}, pages 10619--10629, Online and Punta Cana,
  Dominican Republic. Association for Computational Linguistics.

\bibitem[{Liu et~al.(2018)Liu, Chen, Deng, Zeng, Chen, Li, and
  Tang}]{liu2018lcqmc}
Xin Liu, Qingcai Chen, Chong Deng, Huajun Zeng, Jing Chen, Dongfang Li, and
  Buzhou Tang. 2018.
\newblock Lcqmc: A large-scale chinese question matching corpus.
\newblock In \emph{Proceedings of the 27th International Conference on
  Computational Linguistics}, pages 1952--1962.

\bibitem[{Liu et~al.(2019)Liu, Ott, Goyal, Du, Joshi, Chen, Levy, Lewis,
  Zettlemoyer, and Stoyanov}]{liu2019roberta}
Yinhan Liu, Myle Ott, Naman Goyal, Jingfei Du, Mandar Joshi, Danqi Chen, Omer
  Levy, Mike Lewis, Luke Zettlemoyer, and Veselin Stoyanov. 2019.
\newblock Roberta: A robustly optimized bert pretraining approach.
\newblock \emph{arXiv preprint arXiv:1907.11692}.

\bibitem[{Mirzadeh et~al.(2020)Mirzadeh, Farajtabar, Li, Levine, Matsukawa, and
  Ghasemzadeh}]{mirzadeh2020improved}
Seyed~Iman Mirzadeh, Mehrdad Farajtabar, Ang Li, Nir Levine, Akihiro Matsukawa,
  and Hassan Ghasemzadeh. 2020.
\newblock Improved knowledge distillation via teacher assistant.
\newblock In \emph{Proceedings of the AAAI conference on artificial
  intelligence}, volume~34, pages 5191--5198.

\bibitem[{Rajpurkar et~al.(2018)Rajpurkar, Jia, and
  Liang}]{rajpurkar-etal-2018-know}
Pranav Rajpurkar, Robin Jia, and Percy Liang. 2018.
\newblock \href {https://doi.org/10.18653/v1/P18-2124} {Know what you don{'}t
  know: Unanswerable questions for {SQ}u{AD}}.
\newblock In \emph{Proceedings of the 56th Annual Meeting of the Association
  for Computational Linguistics (Volume 2: Short Papers)}, pages 784--789,
  Melbourne, Australia. Association for Computational Linguistics.

\bibitem[{Sanh et~al.(2019)Sanh, Debut, Chaumond, and Wolf}]{distillbert}
Victor Sanh, Lysandre Debut, Julien Chaumond, and Thomas Wolf. 2019.
\newblock \href {https://doi.org/10.48550/ARXIV.1910.01108} {Distilbert, a
  distilled version of bert: smaller, faster, cheaper and lighter}.

\bibitem[{Shao et~al.(2018)Shao, Liu, Lai, Tseng, and Tsai}]{shao2018drcd}
Chih~Chieh Shao, Trois Liu, Yuting Lai, Yiying Tseng, and Sam Tsai. 2018.
\newblock Drcd: a chinese machine reading comprehension dataset.
\newblock \emph{arXiv preprint arXiv:1806.00920}.

\bibitem[{Shen et~al.(2020)Shen, Dong, Ye, Ma, Yao, Gholami, Mahoney, and
  Keutzer}]{shen2020q}
Sheng Shen, Zhen Dong, Jiayu Ye, Linjian Ma, Zhewei Yao, Amir Gholami,
  Michael~W Mahoney, and Kurt Keutzer. 2020.
\newblock Q-bert: Hessian based ultra low precision quantization of bert.
\newblock In \emph{Proceedings of the AAAI Conference on Artificial
  Intelligence}, volume~34, pages 8815--8821.

\bibitem[{Sun et~al.(2020)Sun, Yu, Song, Liu, Yang, and
  Zhou}]{sun-etal-2020-mobilebert}
Zhiqing Sun, Hongkun Yu, Xiaodan Song, Renjie Liu, Yiming Yang, and Denny Zhou.
  2020.
\newblock \href {https://doi.org/10.18653/v1/2020.acl-main.195}
  {{M}obile{BERT}: a compact task-agnostic {BERT} for resource-limited
  devices}.
\newblock In \emph{Proceedings of the 58th Annual Meeting of the Association
  for Computational Linguistics}, pages 2158--2170, Online. Association for
  Computational Linguistics.

\bibitem[{Tan and Zhang(2008)}]{tan2008empirical}
Songbo Tan and Jin Zhang. 2008.
\newblock An empirical study of sentiment analysis for chinese documents.
\newblock \emph{Expert Systems with applications}, 34(4):2622--2629.

\bibitem[{Vaswani et~al.(2017)Vaswani, Shazeer, Parmar, Uszkoreit, Jones,
  Gomez, Kaiser, and Polosukhin}]{vaswani2017attention}
Ashish Vaswani, Noam Shazeer, Niki Parmar, Jakob Uszkoreit, Llion Jones,
  Aidan~N Gomez, {\L}ukasz Kaiser, and Illia Polosukhin. 2017.
\newblock Attention is all you need.
\newblock \emph{Advances in neural information processing systems}, 30.

\bibitem[{Xia et~al.(2022)Xia, Zhong, and Chen}]{xia2022structured}
Mengzhou Xia, Zexuan Zhong, and Danqi Chen. 2022.
\newblock Structured pruning learns compact and accurate models.
\newblock In \emph{Association for Computational Linguistics (ACL)}.

\bibitem[{Xu et~al.(2020)Xu, Hu, Zhang, Li, Cao, Li, Xu, Sun, Yu, Yu
  et~al.}]{xu2020clue}
Liang Xu, Hai Hu, Xuanwei Zhang, Lu~Li, Chenjie Cao, Yudong Li, Yechen Xu, Kai
  Sun, Dian Yu, Cong Yu, et~al. 2020.
\newblock Clue: A chinese language understanding evaluation benchmark.
\newblock \emph{arXiv preprint arXiv:2004.05986}.

\bibitem[{Yang et~al.(2019)Yang, Dai, Yang, Carbonell, Salakhutdinov, and
  Le}]{yang2019xlnet}
Zhilin Yang, Zihang Dai, Yiming Yang, Jaime Carbonell, Russ~R Salakhutdinov,
  and Quoc~V Le. 2019.
\newblock Xlnet: Generalized autoregressive pretraining for language
  understanding.
\newblock \emph{Advances in neural information processing systems}, 32.

\bibitem[{Yang et~al.(2020)Yang, Cui, Chen, Che, Liu, Wang, and
  Hu}]{yang-etal-2020-textbrewer}
Ziqing Yang, Yiming Cui, Zhipeng Chen, Wanxiang Che, Ting Liu, Shijin Wang, and
  Guoping Hu. 2020.
\newblock \href {https://doi.org/10.18653/v1/2020.acl-demos.2} {{T}ext{B}rewer:
  {A}n {O}pen-{S}ource {K}nowledge {D}istillation {T}oolkit for {N}atural
  {L}anguage {P}rocessing}.
\newblock In \emph{Proceedings of the 58th Annual Meeting of the Association
  for Computational Linguistics: System Demonstrations}, pages 9--16, Online.
  Association for Computational Linguistics.

\end{thebibliography}
\bibliographystyle{acl_natbib}

\end{document}